\documentclass[10pt,twocolumn,letterpaper]{article}

\usepackage[pagenumbers]{cvpr} %

\usepackage{graphicx}
\usepackage{amsmath}
\usepackage{amssymb}
\usepackage{booktabs}

\usepackage[title,toc,titletoc,page]{appendix} %

\usepackage{epsfig}
\usepackage{graphicx}
\usepackage{comment}
\usepackage{amsmath,amssymb} %
\usepackage{xcolor}
\usepackage[export]{adjustbox}
\usepackage{bbm}

\usepackage[font=small,skip=5pt]{caption}
\usepackage{float}

\usepackage{multirow}
\usepackage{tabularx}
\usepackage{soul}
\usepackage{array}
\usepackage{comment}
\usepackage{outlines}

\usepackage[mathscr]{euscript}
\newcolumntype{L}[1]{>{\raggedright\let\newline\\arraybackslash\hspace{0pt}}m{#1}}
\newcolumntype{C}[1]{>{\centering\let\newline\\arraybackslash\hspace{0pt}}m{#1}}
\newcolumntype{R}[1]{>{\raggedleft\let\newline\\arraybackslash\hspace{0pt}}m{#1}}
\newcolumntype{Y}{>{\centering\arraybackslash}X}
\usepackage{booktabs}
\usepackage{enumitem}

\usepackage[pagebackref,breaklinks,bookmarks=false,colorlinks]{hyperref}
\usepackage{amsmath}

\newcommand{\mat}[1]{\mathbf{#1}}

\newcommand{\vect}[1]{\mathbf{#1}}

\newcommand{\pose}[0]{\boldsymbol{\theta}}
\newcommand{\shape}[0]{\boldsymbol{\beta}}

\newcommand{\smpl}[0]{H}

\newcommand{\sifNet}{SIF-Net} %
\newcommand{\hvopNet}{HVOP-Net} %

\newcommand{\ourSMPL}{SMPL-T} %

\newcommand{\poseSeq}{\mat{\Theta}}
\newcommand{\shapeSeq}{\mat{\mathcal{B}}}
\newcommand{\rotSeq}{\mat{\mathcal{R}}^o}
\newcommand{\transSeq}{\mat{\mathcal{T}}^o}
\newcommand{\paramAll}{\mat{\Phi}} %

\usepackage{multirow}

\usepackage[pagebackref,breaklinks,colorlinks]{hyperref}
\usepackage[accsupp]{axessibility} %

\usepackage[capitalize]{cleveref}
\crefname{section}{Sec.}{Secs.}
\Crefname{section}{Section}{Sections}
\Crefname{table}{Table}{Tables}
\crefname{table}{Tab.}{Tabs.}

\def\cvprPaperID{5182} %
\def\confName{CVPR}
\def\confYear{2023}

\begin{document}

\twocolumn[{%
\renewcommand\twocolumn[1][]{#1}%

\title{Visibility Aware Human-Object Interaction Tracking from Single RGB Camera}

\author{
Xianghui Xie 
\qquad\qquad
Bharat Lal Bhatnagar
\qquad\qquad
Gerard Pons-Moll \\
\\
University of T\"ubingen, T\"ubingen AI Center, Germany \\
Max Planck Institute for Informatics, Saarland Informatics Campus, Germany \\
{\tt\small \{xxie, bbhatnag\}@mpi-inf.mpg.de, gerard.pons-moll@uni-tuebingen.de}
}

\maketitle

}]

\begin{abstract}
\textit{Capturing the interactions between humans and their environment in 3D is important for many applications in robotics, graphics, and vision. Recent works to reconstruct the 3D human and object from a single RGB image do not have consistent relative translation across frames because they assume a fixed depth. Moreover, their performance drops significantly when the object is occluded.
In this work, we propose a novel method to track the 3D human, object, contacts between them, and their relative translation across frames from a single RGB camera, while being robust to heavy occlusions. Our method is built on two key insights. First, we condition our neural field reconstructions for human and object on per-frame SMPL model estimates obtained by pre-fitting SMPL to a video sequence. This improves neural reconstruction accuracy and produces coherent relative translation across frames. Second, human and object motion from visible frames provides valuable information to infer the occluded object. We propose a novel transformer-based neural network that explicitly uses object visibility and human motion to leverage neighbouring frames to make predictions for the occluded frames. Building on these insights, our method is able to track both human and object robustly even under occlusions. Experiments on two datasets show that our method significantly improves over the state-of-the-art methods. Our code and pretrained models are available at: \href{https://virtualhumans.mpi-inf.mpg.de/VisTracker}{https://virtualhumans.mpi-inf.mpg.de/VisTracker}.}

\end{abstract}

\section{Introduction}
\begin{figure}[t]
    \centering
    \includegraphics[width=0.47\textwidth]{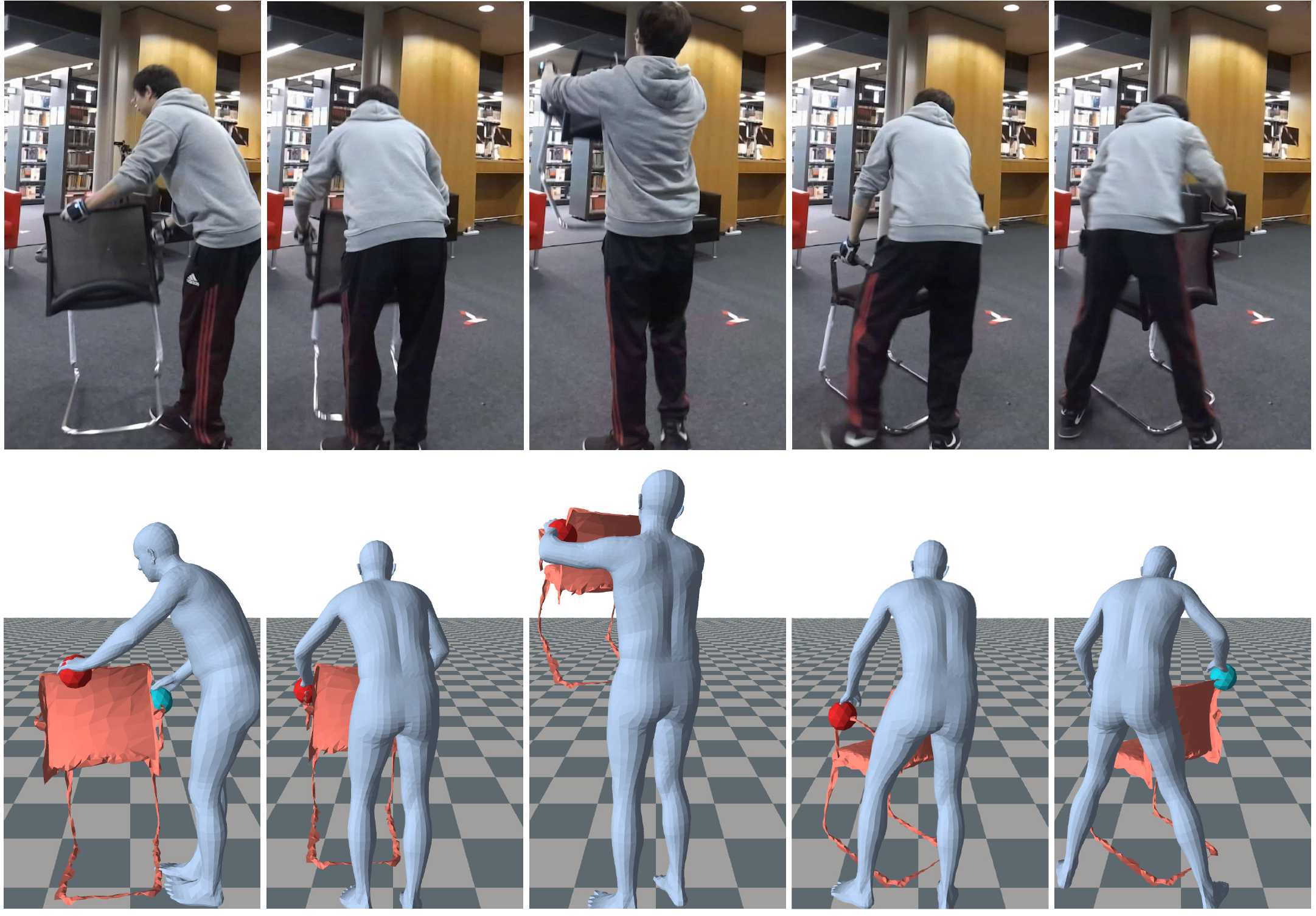}
    \caption{From a monocular RGB video, our method tracks the human, object and contacts between them even under occlusions.}
    \label{fig:teaser}
\end{figure}

Perceiving and understanding human as well as their interaction with the surroundings has lots of applications in robotics, gaming, animation and virtual reality etc.
Accurate interaction capture is however very hard. 
Early works employ high-end systems such as dense camera arrays \cite{cvpr18total-capture, SIGGRAH08garment-capture, TG15free-viewpoint-video} that allow accurate capture but are expensive to deploy. Recent works \cite{bhatnagar22behave, jiang2022neuralfusion, huang2022rich} reduce the requirement to multi-view RGBD cameras but it is still complicated to setup the full capture system hence is not friendly for consumer-level usage. This calls for methods that can capture human-object interaction from a single RGB camera, which is more convenient and user-friendly. 

However, reasoning about the 3D human and object from monocular RGB images is very challenging.
The lack of depth information makes the predictions susceptible to depth-scale ambiguity, leading to temporally incoherent tracking. Furthermore, the object or human can get heavily occluded, making inference very hard. 
Prior work PHOSA~\cite{zhang2020phosa} relies on hand-crafted heuristics to reduce the ambiguity but such heuristic-based method is neither very accurate nor scalable. More recently, CHORE \cite{xie22chore} combines neural field reconstructions with model based fitting obtaining promising results.  
However, CHORE, assumes humans are at a fixed depth from the camera and predicts scale alone, thereby losing the important \emph{relative translation} across frames. 
Another limitation of CHORE is that it is not robust under occlusions as little information is available from single-frame when the object is barely visible. Hence CHORE often fails in these cases, see \cref{fig:main-result}. 

In this work, we propose the first method that can track both human and object accurately from monocular RGB videos. Our approach combines neural field predictions and model fitting, which has been consistently shown to be more effective than directly regressing pose~\cite{bhatnagar2020ipnet,bhatnagar22behave,xie22chore,bhatnagar2020loopreg}.
In contrast to existing neural field based reconstruction methods~\cite{pifuSHNMKL19,xie22chore}, we can do tracking including inference of relative translation.
Instead of assuming a fixed depth, we condition the neural field reconstructions (for object and human) on per frame SMPL estimates (\ourSMPL{}) including translation in camera space obtained by pre-fitting SMPL to the video sequence. This results in coherent translation and improved neural reconstruction. 
In addition, we argue that during human-object interaction, the object motion is highly correlated with the human motion, which provides us valuable information to recover the object pose even when it is occluded (see \cref{fig:teaser} column 3-4). 
To this end, we propose a novel transformer based network that leverages the human motion and object motion from nearby visible frames to predict the object pose under heavy occlusions. 

We evaluate our method on the BEHAVE \cite{bhatnagar22behave} and InterCap dataset~\cite{huang2022intercap}. Experiments show that our method can robustly track human, object and realistic contacts between them even under heavy occlusions and significantly outperforms the currrent state of the art method, CHORE~\cite{xie22chore}. We further ablate the proposed \emph{\ourSMPL{} conditioning} and \emph{human and visibility aware} object pose prediction network and demonstrate that they are key for accurate human-object interaction tracking.

\noindent In summary, our key contributions include:
\begin{itemize}
    \itemsep0em %
    \item We propose the first method that can jointly track full-body human interacting with a movable object from a monocular RGB camera. 
    \item We propose \emph{\ourSMPL{} conditioned interaction fields}, predicted by a neural network that allows consistent 4D tracking of human and object.
    \item We introduce a novel \emph{human and visibility aware} object pose prediction network along with an object visibility prediction network that can recover object poses even under heavy occlusions. 
    \item Our code and pretrained models are publicly available to foster future research in this direction.  
\end{itemize}

\begin{figure*}
    \centering
    \fbox{\includegraphics[width=0.98\textwidth]{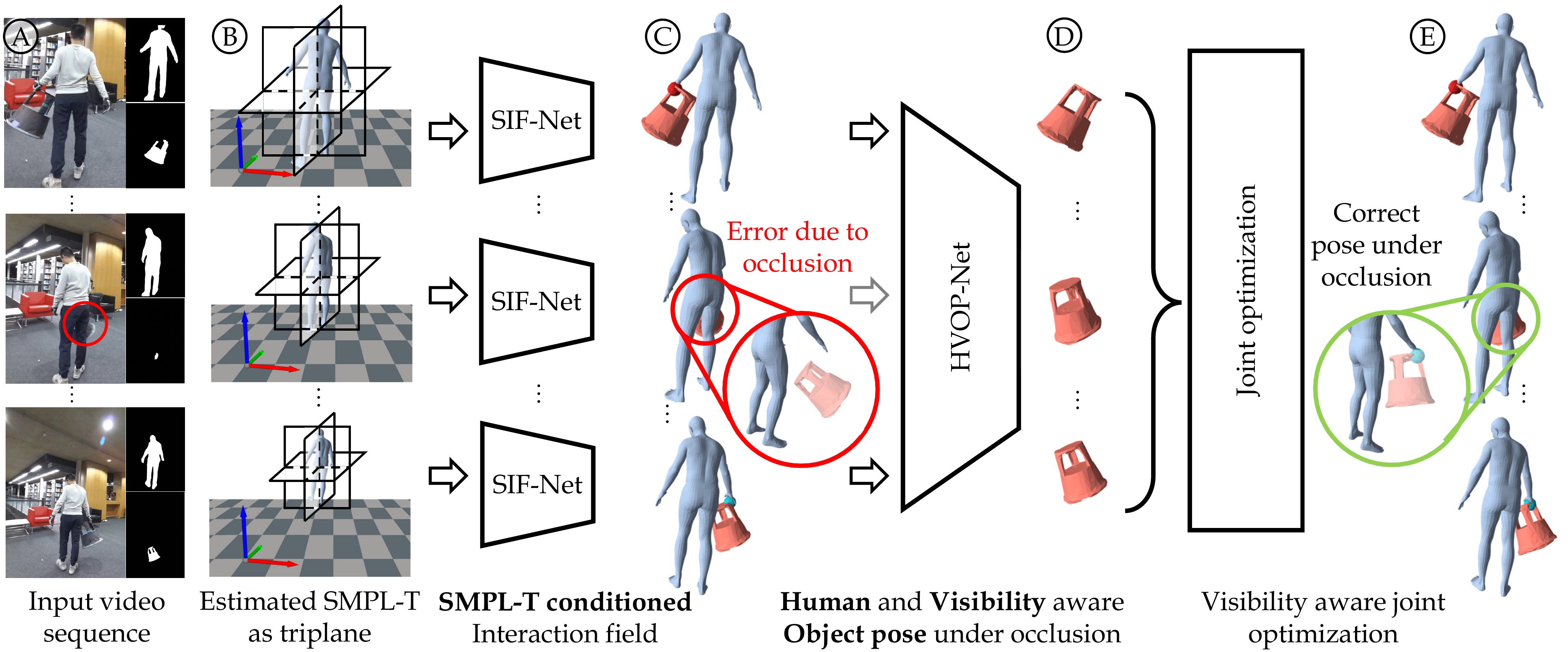}}
    \caption{Given an input RGB sequence of a human interacting with an object and their corresponding human-object masks (A), we aim to reconstruct and track the 3D human, object and the contacts between them (E). Our first key idea is a \ourSMPL{} conditioned interaction field network (SIF-Net, details in \cref{sec:smpl-interaction-field}) that predicts neural fields conditioned on estimated SMPL meshes in camera space (col. B, \ourSMPL{}, details in \cref{sec:TranSMPL}). 
    \ourSMPL{} conditioning provides us temporally consistent relative translation, which is important for coherent 4D tracking. Our second key insight is to predict object pose under occlusions (D) leveraging human motion and object visibility information (HVOP-Net, details in \cref{sec:object-motion-infill}). This prediction provides robust object tracking for frames with heavy occlusion. We then jointly optimize human and object (details in ~\cref{sec:joint-optimization}) to satisfy image observations and contact constraints. }
    
    \label{fig:method}
\end{figure*}

\section{Related Work}
In this section, we first review recent works that deal with human or object pose estimation and tracking separately. We then discuss recent progresses that model human and object interactions and works that deal with occlusions explicitly. 

\textbf{Human or object pose estimation and tracking.} 
After the introduction of SMPL~\cite{smpl2015loper} body model, tremendous progress has been made in human mesh recovery (HMR) from single images \cite{PonsModelBased,bogo2016smplify,SMPL-X:2019,alldieck2018video,alldieck2018detailed,kolotouros2019spin, coronaLVD} or videos \cite{yuan2022glamr, rajasegaran2021tp3d, rajasegaran2022tracking-phalp, kocabas2019vibe, dai2023sloper4d}. We refer readers to a recent review of HMR methods in~\cite{tian2022hmrsurvey}. 
On the other hand, deep learning method has also significantly improved object 6D pose estimation from single RGB images \cite{Wang_2021_GDRN, majcher_shape_nodate, fan_deep_2022, li_deepim_nodate, di_so-pose_2021, peng_pvnet_2019, hu_single-stage_cvpr20}. However, object pose tracking has received less attention and most works focus on RGBD inputs~\cite{wen2020se, stoiber_iterative_2022, zheng_tp-ae_2022, deng_poserbpf_2021, wen_bundletrack_2021}. Two works explore the camera localization ideas from SLAM communities and can track object from RGB videos \cite{liu2022gen6d, sun2022onepose}. Nevertheless, they heavily rely on visual evidence and the performance is unknown under heavy occlusions. They also do not track human-object interactions.

\textbf{Human-object interaction.} 
Modelling human object interaction is an emerging research topic in recent years. Hand-object interaction is studied with works modelling hand-object interaction from RGB~\cite{GrapingField:3DV:2020,Corona_2020_CVPR,hasson19_obman, ehsani2020force, yang2021cpf}, RGBD~\cite{Brahmbhatt_2019_CVPR,Brahmbhatt_2020_ECCV, hampali2020honnotate} or 3D inputs~\cite{GRAB:2020,ContactGrasp2019Brahmbhatt, zhou2022toch, petrov2020objectpopup}. There are also works that model human interacting with a static scene \cite{PROX:2019, huang2022rich, shimada2022hulc, weng2020holistic, Yi_MOVER_2022, CVPR21HPS, yi2022mime} or deformable surface \cite{Li_3DV2022MocapDeform}. More recently, the release of BEHAVE\cite{bhatnagar22behave} and InterCap\cite{huang2022intercap} datasets allows bench-marking of full-body interacting with a movable object. However, human-object interaction capture usually deploys multi-view RGB \cite{sun2021HOI-FVV} or RGBD\cite{bhatnagar22behave, huang2022rich, huang2022intercap, zhang2022couch, hampali2020honnotate, Brahmbhatt_2019_CVPR, hampali2020honnotate, dai2023sloper4d} cameras. Only a few works \cite{zhang2020phosa,  xie22chore, wang2022reconstruction} reconstruct dynamic human and object from monocular RGB input and our experiments show that they are not suitable for tracking.

\textbf{Pose estimation under occlusion.} 
Most existing methods assume occlusion-free input images hence are not robust under occlusions. Only a few methods address human pose estimation under partial occlusions ~\cite{Kocabas_PARE_2021, ZhangCVPR20OOH, rempe2021humor, Rockwell2020partial-occlusion, Fieraru3DHumanHumanInteraction, Kocabas_SPEC_2021} or long term occlusions \cite{yuan2022glamr}. For object pose estimation, pixel-wise voting \cite{peng_pvnet_2019} and self-occlusion \cite{di_so-pose_2021} are explored for more robust prediction under occlusions. More recently,  TP-AE~\cite{zheng_tp-ae_2022} predicts object occlusion ratio to guide pose estimation but relies on depth input. Although being impressive on separate human or object pose estimation, these methods do not reason about human-object interaction. Our method is the first one that takes both human and object visibility into account for interaction tracking.

\section{Method}

We present a novel method for jointly tracking the human, the object and the contacts between them, in 3D, from a monocular RGB video.
The first main challenge in monocular tracking is the estimation of human and object translations in camera space due to the depth-scale ambiguity problem. 
Existing method, CHORE~\cite{xie22chore}, reconstructs human and object at a fixed depth to the camera, leading to inconsistent 3D translation across frames. Our key idea is to fit a SMPL model with single shape parameters to a video sequence to obtain consistent relative translation across frames. We call the estimated SMPL as \ourSMPL{} and describe it in more details in \cref{sec:TranSMPL}. Based on the estimated \ourSMPL{}, we then jointly model the 3D human, object and the interactions using our proposed \emph{\ourSMPL{} conditioned Interaction Fields} network (\sifNet{}, \cref{sec:smpl-interaction-field})

Object tracking from a single-frame only, is difficult when the object is barely visible. Hence we introduce a \emph{Human and Visibility aware Object Pose Network} (\hvopNet{}) that leverages human and object motion from visible frames to recover the object under occlusion (\cref{sec:object-motion-infill}). We then use the \sifNet{} and \hvopNet{}  outputs to optimize SMPL model and object pose parameters of this sequence to satisfy neural prediction and image observations (\cref{sec:joint-optimization}). An overview of our approach can be found in~\cref{fig:method}.

\subsection{Preliminaries} \label{sec:preliminary}
In this work, we focus on a single human interacting with an object, which is a common setting in other hand-object interaction \cite{yang2021cpf, GRAB:2020, zhou2022toch} and full body-object interaction works \cite{xie22chore, bhatnagar22behave, huang2022intercap}. We represent human using the SMPL \cite{smpl2015loper} body model $\smpl(\pose, \shape)$ that parameterises the 3D human mesh using pose $\pose$ (including global translation) and shape $\shape$ parameters. The object is represented by a known mesh template and we estimate the rotation $\mathbf{R}^o\in SO(3)$ and translation $\vect{t}^o\in \mathbb{R}^3$ parameters. Given a video sequence $\{\mathbf{I}_1, ..., \mathbf{I}_T\}$ where $\mathbf{I}_i \in \mathbb{R}^{H\times W \times 5}$ (RGB, human and object masks), our goal is to estimate the SMPL shape $\shape$, a sequence of SMPL pose $\poseSeq=\{\pose_1, ..., \pose_T\}$, object rotation $\rotSeq=\{\mathbf{R}_1^o, ..., \mathbf{R}_T^o\}$ and translation $\transSeq=\{\vect{t}^o_1, ..., \vect{t}^o_T\}$ parameters that satisfy 2D image observation and realistic interaction constraints between the human and the object.

\subsection{SMPL-T: Temporally consistent SMPL meshes in camera space} 
\label{sec:TranSMPL}
Our first step is to obtain SMPL meshes in camera space that have consistent translation in a video sequence. We leverage 2D body keypoint predictions from openpose\cite{PAMI19openpose} and natural motion smoothness for this. Specifically, we use FrankMocap~\cite{rong2020frankmocap} to initialise SMPL-T pose $\poseSeq=\{\pose_1,...,\pose_T\}$ and shape $\shapeSeq=\{\shape_1, ..., \shape_T\}$ parameters for a sequence of images. Note here the original SMPL meshes from FrankMocap are centered at origin. We average the SMPL-T shape parameters over the sequence as the shape for the person and optimize the SMPL-T global translation and body poses to minimize 2D reprojection and temporal 
smoothness error: 
\begin{equation}
    E(\poseSeq) = \lambda_\text{J2D}L_\text{J2D} + \lambda_\text{reg}L_\text{reg} + \lambda_a L_{\text{accel}} + \lambda_\text{pi}L_\text{pi}
    \label{eq:objective_smpl_fit}
\end{equation}
where $L_\text{J2D}$ is the sum of body keypoint reprojection losses\cite{bogo2016smplify} over all frames and $L_\text{reg}$ is a regularization on body poses using priors learned from data\cite{smpl2015loper, SMPL-X:2019, tiwari22posendf}. 
$L_{\text{accel}}$ is a temporal smoothness term that penalizes large accelerations over SMPL-T vertices $H_i$: $L_{\text{accel}}=\sum_{i=0}^{T-2}||H_i-2H_{i+1}+H_{i+2}||_2^2$. $L_\text{pi}$, an L2 loss between optimized and initial body pose, prevents the pose from deviating too much from initialization. $\lambda_*$ denotes the loss weights detailed in Supp.

Note that this optimization does not guarantee that we get the absolute translation in the world coordinates, it just ensures that our predictions will be consistent with the SMPL model over a sequence, i.e. we will be off by one rigid transformation for the entire sequence. 

\subsection{\sifNet{}: \ourSMPL{} conditioned interaction field}\label{sec:smpl-interaction-field}
Our \ourSMPL{} provides translation about the human but does not reason about the object and the interaction between them.
Existing method CHORE~\cite{xie22chore} can jointly reason human and object but their humans are predicted at fixed depth. Our key idea is to leverage our \ourSMPL{} meshes to jointly reason human, object and interaction while having consistent relative translation. We model this using a single neural network  which we call \sifNet{}.
The input to \sifNet{} consists of RGB image, human and object masks, and our estimated \ourSMPL{}. With features from \ourSMPL{} and input images, it then predicts interaction field which consists of human and object distance fields, SMPL part correspondence field, object pose and visibility field. 

\textbf{\sifNet{} feature encoding.} 
Existing neural implicit methods \cite{saito2020pifuhd, xie22chore, pifuSHNMKL19} rely mainly on features from input image, a main reason that limits their human prediction at fixed depth. Instead, we extract features from both our estimated \ourSMPL{} meshes and input image, providing more distinct features for the query points along the same ray. 
Inspired by EG3D~\cite{Chan2022EG3D}, we use the triplane representation for \ourSMPL{} feature learning due to its efficiency. 
Specifically, we use orthographic camera $\pi^o(\cdot)$, to render the \ourSMPL{} mesh silhouette from right, back and top-down views and obtain three images $\mathbf{S}^r_i, \mathbf{S}^b_i, \mathbf{S}^t_i$ respectively, where $\mathbf{S}^i\in \mathbb{R}^{H\times W}$, see supplementary for more visualization. Note here the triplane origin is placed at our \ourSMPL{} mesh center (\cref{fig:method} B).
We then train an image encoder $f^\text{tri}(\cdot)$ that extracts a pixel aligned feature grid $\mat{D}^j_i\in \mathbb{R}^{H_c\times W_c \times C}$ from each rendered view $\mathbf{S}^j_i$, where $j\in \{r, b, t\}$ and $H_c, W_c, C$ are the feature grid dimensions.
To extract features for a query point $\vect{p} \in \mathbb{R}^3$, we project $\vect{p}$ into the three planes using the same orthographic projection $\pi_\vect{p}^o=\pi^o(\vect{p})$ and extract local features $ \mat{D}_i^{\vect{p}} = (\mat{D}^r_i(\pi_\vect{p}^o), \mat{D}^b_i(\pi_\vect{p}^o), \mat{D}^t_i(\pi_\vect{p}^o))$ using bilinear interpolation.

In addition to the \ourSMPL{} features, \sifNet{} also extracts information from input images. More specifically, we train an image encoder $f^\text{enc}(\cdot)$ to extract feature grid $\mat{Z}_i \in \mathbb{R}^{H_f\times W_f \times C_f}$ from input image $\mat{I}_i\in \mathbb{R}^{H \times W\times 5}$, here $H_f, W_f, C_f$ and $H, W$ are feature grid and input image dimensions respectively. Given query point $\vect{p}\in \mathbb{R}^3$, we project it to 2D image using full perspective projection $\pi_\vect{p} = \pi(\vect{p})$ and extract pixel-aligned features $\mat{Z}_i^\vect{p} = \mat{Z}_i(\pi_\vect{p})$. The input image feature is concatenated with the \ourSMPL{} feature to form an input and translation aware point feature: $\mat{F}_i^\vect{p}=(\mat{Z}_i^\vect{p}, \mat{D}_i^\vect{p})$.

\textbf{\sifNet{} predictions.}
From the point feature $\mat{F}^\vect{p}_i$ discussed above, we predict our interaction fields that jointly model human, object and their contacts, similar to CHORE~\cite{xie22chore}. Specifically, we predict the unsigned distances to human and object surfaces using $f^u:\mat{F}_i^\vect{p}\mapsto \mathbb{R}^2_{\geq0}$. This allows fitting the SMPL mesh and object template by minimizing the predicted distances at mesh vertices.
We can also infer contacts, as the points having small distances to both human and object surfaces.
For more robust SMPL fitting~\cite{bhatnagar2020ipnet} and modelling which body part the object point is in contact with, we predict SMPL part correspondence using $f^p: \mat{F}_i^\vect{p}\mapsto \{1,2,...,K\}$ where $K$ is the number of SMPL parts. For more accurate object fitting, we additionally predict object rotation with $f^R:\mat{F}_i^\vect{p}\mapsto \mathbb{R}^{3\times 3}$ and translation with $f^c: \mat{F}_i^\vect{p}\mapsto \mathbb{R}^3$. The predicted $3\times 3$ matrix is projected to SO(3) using symmetric orthogonalization\cite{levinson20nisp_so3_rot}. At test time, we first use $f^u$ to find points on the object surface\cite{chibane2020ndf} and take the average rotation and translation predictions of these points as the object pose. To handle occlusions, we also predict the object visibility field using $f^\text{vis}:\mat{F}_i^\vect{p}\mapsto [0, 1]$. The visibility is useful to recover the object pose under occlusion, see more details in \cref{sec:object-motion-infill}. 

\textbf{Why use triplane to encode \ourSMPL{}?} A direct alternative to triplane based \ourSMPL{} encoding is to find the closest point in \ourSMPL{} mesh and concatenate that coordinate to the point features. But such a method is slow (computing point to surface distance) and does not allow flexible learning of local and global features. Another choice is to voxelize the \ourSMPL{} mesh and extract point local features using IF-Nets~\cite{chibane20ifnet} but such a method is still expensive. Therefore we chose the more efficient triplane representation to encode our estimated \ourSMPL{} meshes.

\textbf{Implementation.} Our \ourSMPL{} feature extractor $f^\text{tri}$ is shared for three views and trained end to end with image encoder $f^\text{enc}$ and other neural field predictors. At training time, we input the renderings from ground truth SMPL meshes and train the network to predict GT labels. At test time, we obtain the \ourSMPL{} meshes using \cref{eq:objective_smpl_fit}. In order to have smoother \ourSMPL{} feature in a sequence, we use SmoothNet \cite{zeng2022smoothnet} to smooth the optimized SMPL parameters. We evaluate this component and provide more implementation details in supplementary. 

\subsection{\hvopNet{}: Human and Visibility aware Object Pose under occlusions}\label{sec:object-motion-infill}
Our \sifNet{} recovers translation and more accurate object pose. However, the object pose prediction under very heavy occlusions remains challenging because no image evidence from single frame is available for accurate prediction, see \cref{fig:eval-pose-pred}. Our key idea is to use the \ourSMPL{} and object pose
from other visible frames to predict the object of occluded frames. 
To this end, we first predict object visibility scores in each image, which are then leveraged together with the human evidence from neighbouring frames to predict the object poses of the occluded frames.

\textbf{Object visibility score.} Our visibility score denotes how much the object is visible in the input image. We train a visibility decoder $f^\text{vis}(\cdot)$, a prediction head of \sifNet{}, that takes a point feature $\mat{F}_i^\vect{p}$ as input and predicts visibility score $v_i\in[0, 1]$ for frame $i$. At test time, we first use the neural object distance predictor $f^u$ to find object surface points \cite{chibane2020ndf} and then take the average visibility predictions of these points as the object visibility score for this image. 

\textbf{Object pose prediction under heavy occlusion.}
Our goal now is to predict accurate object pose for heavily occluded frames. We consider frames whose visibility score $v_i$ is smaller than $\delta=0.5$ as the occluded frames.
Inspired by works from motion infill \cite{KIM2022108894CMIB, Duan_unified-motion-aaai22} and synthesis \cite{yi2022generating, zhang2022couch}, we design our \hvopNet{} that leverages transformer ~\cite{NIPS2017_attention} and explicitly takes the human motion and object visibility into account to recover object pose under heavy occlusions. 

More specifically, we first use a transformer $f^s(\cdot)$ to aggregate temporal information of the \ourSMPL{} poses: $f^s:\mathbb{R}^{T\times |\theta_i|}\mapsto \mathbb{R}^{T\times D_{hs}}$, where $|\theta_i|$ is the SMPL pose dimension and $D_{hs}$ is the hidden feature dimension. Similarly, we use a transformer $f^o(\cdot)$ to aggregate temporal information of the object poses: $f^o:\mathbb{R}^{T\times D_o}\mapsto \mathbb{R}^{T\times D_{ho}}$. Note here the \ourSMPL{} transformer $f^s$ attends to all frames while the object transformer $f^o$ only attends to frames where object is visible ($v_i\geq\delta$). We then concatenate the \ourSMPL{} and object features and use another transformer $f^\text{comb}$ to aggregate both human and object information and predict the object poses: $f^\text{comb}: \mathbb{R}^{T\times (D_{hs} + D_{ho})}\mapsto \mathbb{R}^{T\times D_{o}}$. 
The joint transformer $f^\text{comb}$ attends to all frames. Our experiments show that our \hvopNet{} is important to accurately predict object pose under heavy occlusions, see \cref{fig:eval-pose-pred}.

\textbf{Implementation.} The visibility decoder $f^\text{vis}$ is trained end to end with other \sifNet{} components using L2 loss. The GT visibility score is computed as the number of visible object pixels (from object mask) divided by total number of object pixels (from GT object rendering). To train our \hvopNet{}, we randomly zero out the object pose for a small clip of the input sequence and provide ground truth SMPL and object poses for other frames as input. We train our network to accept input sequence of a fixed length but at test time, the sequence can have various length and object occlusion can last for a longer time. To this end, we use an auto-regressive algorithm to recover the object pose of a full video, similar to \cite{yuan2022glamr}. At test time, we use \sifNet{} object pose predictions and zero out highly occluded frames based on the predicted visibility scores. We empirically find that having smooth object pose as input is helpful for more accurate prediction. Hence we use SmoothNet~\cite{zeng2022smoothnet} to smooth \sifNet{} object pose predictions before inputing them to \hvopNet{}. The evaluation of this component and more training details are described in our supplementary.

\subsection{Visibility aware joint optimization}\label{sec:joint-optimization}
To obtain SMPL and object meshes that align with input images and satisfy contact constraints, we leverage our network predictions from \cref{sec:smpl-interaction-field} and \cref{sec:object-motion-infill} to formulate a robust joint optimization objective.
Our goal is to obtain an optimal set of parameters $\paramAll=\{\poseSeq, \shape, \rotSeq, \transSeq\}$ for SMPL pose, shape, object rotation and translation respectively. We initialize the SMPL parameters from our estimated \ourSMPL{} (\cref{eq:objective_smpl_fit}) and object parameters from our \hvopNet{} predictions (\cref{sec:object-motion-infill}). Inspired by CHORE\cite{xie22chore}, our energy function consists of human data $E_{data}^h$, object data $E_{data}^o$, contact data $E_{data}^c$ and SMPL pose prior term $E_{reg}$: 
\begin{equation}
    E_(\paramAll) = E_{data}^h + E_{data}^o + E_{data}^c + E_{reg}.
    \label{eq:objective_joint}
\end{equation}
here $E_{reg}$ is a body pose and shape prior loss\cite{smpl2015loper}. We explain other loss data terms next. 
\\
\noindent
\textbf{Human data term}. $E_{data}^h$ minimizes the discrepancy between \sifNet{} prediction and SMPL meshes as well as a temporal smoothness error: $E_{data}^h(\poseSeq, \shape)=\sum_{i=1}^T L_{\text{neural}}^h(\pose_i, \shape) + \lambda_{\text{ah}} L_{\text{accel}}(\poseSeq)$, where $L_{\text{accel}}$ is the same used in \cref{eq:objective_smpl_fit}. $L_{\text{neural}}^h$ pushes the SMPL vertices to the zero-level set of the human distance field represented by neural predictor $f^{u,h}_i$ and forces correct SMPL part locations predicted by $f^p_i$: 
\begin{multline}
    E^h_\text{neural}(\poseSeq, \shape) = \sum_{i=1}^T(\sum_{\vect{p}\in \smpl(\pose_i, \shape)}(\lambda_{h}\min(f^{u,h}_i(\mat{F}^\vect{p}_i), \delta_h) +\\ \lambda_{p} L_p(l_\vect{p}, f^p_i(\mat{F}^\vect{p}_i))))
\end{multline}
here $l_\vect{p}$ is the predefined SMPL part label \cite{bhatnagar2020ipnet} of SMPL vertex $\vect{p}$ and $L_p$ is the categorical cross entropy loss function. $\delta_h$ is a small clamping value. 
\\
\textbf{Object data term.} We transform the object template vertices $\mat{O}\in \mathbb{R}^{3\times N}$ using object pose parameters of frame $i$ by: $\mat{O}^\prime_i=\mat{R}^o_i\mat{O} + \vect{t}^o_i$. Intuitively, the object vertices should lie on the zero-level set of the object distance field represented by $f^{u,o}_i$ and the rendered silhouette should match the 2D object mask $\mat{M}^o_i$. Hence we formulate the loss as: 
\begin{multline}
E^o_{data}(\rotSeq, \transSeq) = \sum_{i=1}^{T}v_i(\sum_{\vect{p}\in \mat{O}^\prime_i}\lambda_o\min(f^{u,o}_i(\mat{F}^\vect{p}_i), \delta_o)\\ 
    + \lambda_{\text{occ}}L_{\text{occ-sil}}(\mat{O}_i^\prime, \mat{M}^o_i)) + \lambda_{\text{ao}}L_{\text{ao}}
    \label{eq:objective_object_fit}
\end{multline}
where $v_i$ is the predicted object visibility score described in \cref{sec:object-motion-infill}. This down-weights the loss values of network predictions for frames where object is occluded and allows more temporal regularization. $L_\text{occ-sil}$ is an occlusion-aware silhouette loss\cite{zhang2020phosa} and $L_{\text{ao}}$ is a temporal smoothness loss applied to object vertices $\mat{O}^\prime_i$, similar to $L_{\text{accel}}$ in \cref{eq:objective_smpl_fit}. 

\noindent
\textbf{Contact data term.}
The contact data term \cite{xie22chore} minimizes the distance between human and object points that are predicted to be in contact: 
\begin{equation}
    E_\text{data}^c(\rotSeq, \transSeq) = \lambda_c \sum_{i=1}^T(\sum_{j=1}^{K} d(\smpl_j^c(\pose_i, \shape), \mat{O}_{ij}^c))
\end{equation}
here $d(\cdot, \cdot)$ is chamfer distance. We consider human points on the $j^\text{th}$ body part of SMPL mesh $H_i$ of frame $i$ (denoted as $H_{ij}$) are in contact when their predicted distance to the object is smaller than a threshold: $H^c_j(\pose_i, \shape)=\{\vect{p} | \vect{p} \in H_{ij}\, \text{and}\, f^{u, o}_i(\mat{F}^\vect{p}_i)\leq \epsilon\}$. Similarly, we find contact points on object meshes with $\mat{O}_{ij}^c=\{\vect{p} | \vect{p}\in\mat{O}^\prime_i\, \text{and}\, f^{u, h}_i(\mat{F}^\vect{p}_i)\leq\epsilon\, \text{and}\, f^p_i(\mat{F}^\vect{p}_i)=j\}$.  

Please see Supp. for more details about loss weights $\lambda_*$.

\section{Experiments}
\begin{figure*}[t]
    \centering
    \includegraphics[width=\textwidth]{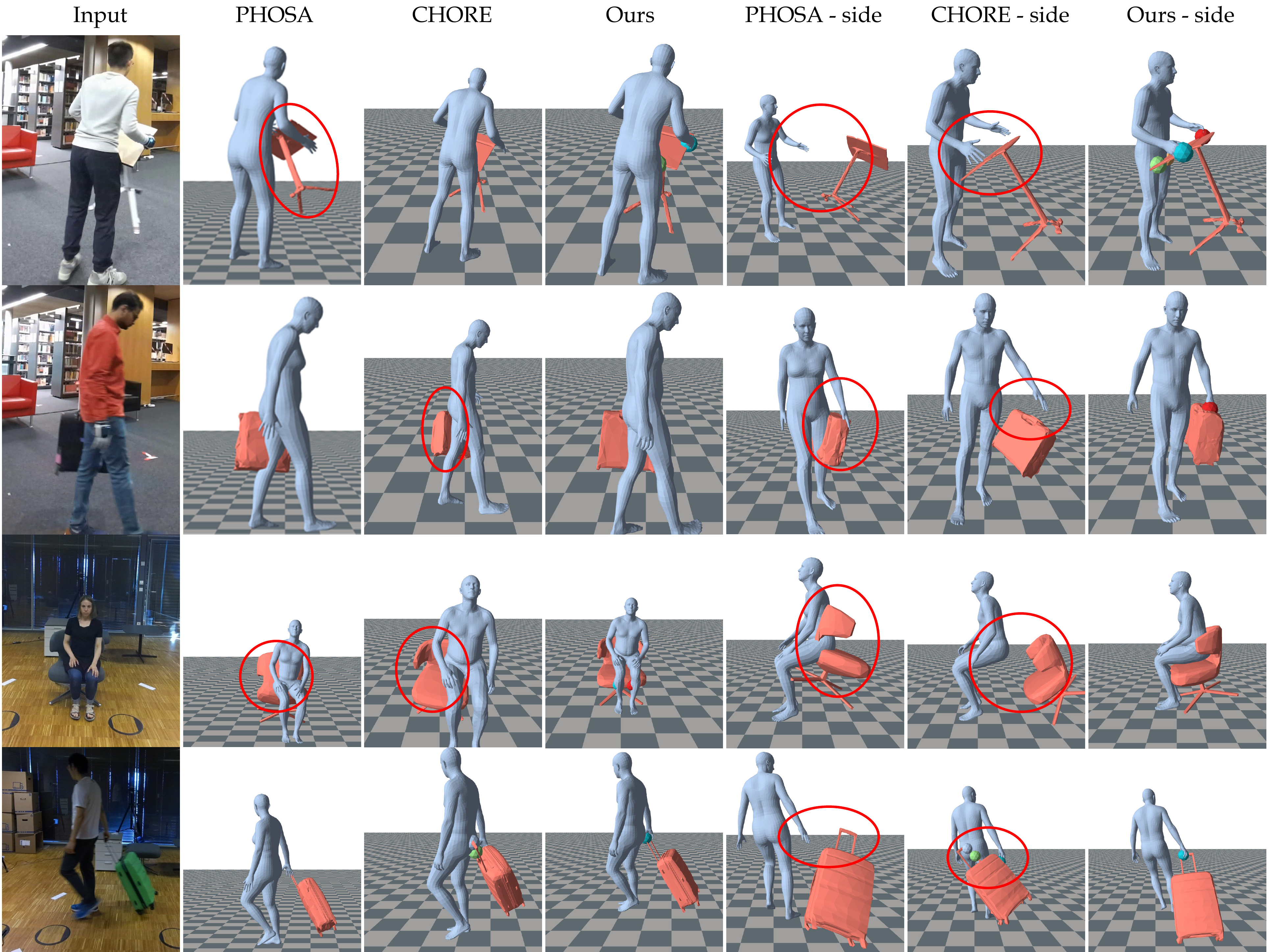}
    
    \caption{Comparison with PHOSA \cite{zhang2020phosa} and CHORE\cite{xie22chore} on BEHAVE\cite{bhatnagar22behave} (row 1-2) and InterCap~\cite{huang2022intercap} (row 3-4). PHOSA's object pose optimization often gets stuck in local minima due to heavy occlusions. CHORE also fails to predict accurate object pose as it does not fully explore the human, temporal and visibility information while our method can robustly track human and object in these challenging cases.}
    \label{fig:main-result}
\end{figure*}

In this section, we first compare our method against existing approaches on tracking human and object and then evaluate the key components of our methods. Our experiments show that our method clearly outperforms existing joint human object reconstruction method and our novel \emph{\ourSMPL{} conditioned interaction fields} (\sifNet{}) as well as \emph{human and visibility aware} object pose prediction (\hvopNet{}) works better than existing state of the art methods. 

\textbf{Baselines.} \textbf{(1) Joint human and object tracking.} We compare against PHOSA \cite{zhang2020phosa} and CHORE~\cite{xie22chore} in the joint human-object reconstruction task.
\textbf{(2) Object pose prediction.} Our \hvopNet{} leverages nearby (un-occluded) frames to predict the object pose of occluded frames. We compare this with a simple baseline that linearly interpolates the object pose between visible frames to recover occluded poses. We also find similarity between our task and motion smoothing/infilling. Hence we compare our \hvopNet{} with SoTA smoothing ~\cite{zeng2022smoothnet} and infill method~\cite{KIM2022108894CMIB}. 

\textbf{Datasets.} We conduct experiments on the BEHAVE~\cite{bhatnagar22behave} and InterCap~\cite{huang2022intercap} dataset. 
\textbf{(1) BEHAVE} ~\cite{bhatnagar22behave} captures 7 subjects interacting with 20 different objects in natural environments and contains SMPL and object registrations annotated at 1fps.
We use the extended BEHAVE dataset, which registers SMPL and object for BEHAVE sequences at 30 fps. We follow the official split~\cite{xie22chore} with $217$ sequences for training and $82$ for testing.
\textbf{(2) InterCap}~\cite{huang2022intercap} is a similar dataset that captures 10 subjects interacting with 10 different objects. The dataset comes with pseudo ground truth SMPL and object registrations at 30fps. We train our model on sequences from subject 01-08 (173 sequences) and test on sequences from subject 09-10 (38 sequences). 

We compare with CHORE~\cite{xie22chore} on the full test set of both datasets. PHOSA~\cite{zhang2020phosa} optimizes hundreds of random object pose initialization per image hence the inference speed is very slow (2min/image), which makes it infeasible to run on full video sequences. Hence we compare with PHOSA on key frames only, denoted as $\text{BEHAVE}^*$(3.9k images) and $\text{InterCap}^*$(1.1k images) respectively.  
Due to the large number of frames in the full BEHAVE test set (127k frames), we conduct other ablation experiments in a sub test set (42k frames) of BEHAVE.

\textbf{Evaluation metrics.}
\textbf{(1) Joint human-object tracking.} We evaluate the performance of SMPL and object reconstruction using Chamfer distance between predicted SMPL and object meshes, and the ground truth. CHORE~\cite{xie22chore} uses Procrustes alignment on combined SMPL and object meshes for \emph{each frame} before computing errors. However, this does not reflect the real accuracy in terms of the relative translation between nearby frames in a video. Inspired by the world space errors proposed by SPEC\cite{Kocabas_SPEC_2021} and GLAMR\cite{yuan2022glamr}, we propose to perform joint Procrustes alignment in a sliding window, as also used in SLAM evaluations \cite{RoNIN-inertial-ICRA20, sturm:hal-rgbd-slam}. More specifically, we combine all SMPL and object vertices within a sliding window and compute a single optimal Procrustes alignment to the ground truth vertices. This alignment is then applied to all SMPL and object vertices within this window and Chamfer distance of SMPL and object meshes are computed respectively. We report both the errors using per-frame alignment (w=1) and alignment with a sliding window of 10s (w=10).
\\
(2) \textbf{Object only evaluation.} For experiments evaluating object pose only, we evaluate the rotation accuracy using rotation angle \cite{object_pose_evaluation}. 
The object translation error is computed as the distance between reconstructed and GT translation. We report all errors in centimetre in our experiments. 

\subsection{Evaluation of tracking results}

\begin{table}[h]
    \footnotesize
    \centering
    \begin{tabular}{l | l | c c c c c}
    \toprule[1.5pt]
    \multirow{2}{*}{Dateset} & \multirow{2}{*}{Methods} & \multicolumn{2}{c}{Align w=1} & \multicolumn{2}{c}{Align w=10} \\
     &  & { SMPL $\downarrow$} & { Obj. $\downarrow$} & { SMPL $\downarrow$} & { Obj. $\downarrow$} \\
    \midrule
    \multirow{2}{*}{BEHAVE} & CHORE & 5.55 & 10.02 & 18.33 & 20.32 \\
    & { Ours} & {\bf 5.25 } & {\bf 8.04} & {\bf 7.81 } & {\bf 8.49}\\
    \midrule
    \multirow{3}{*}{$\text{BEHAVE}^*$} & PHOSA & 12.86 & 26.90 & 27.01 & 59.08 \\
    & CHORE & 5.54 & 10.12 & 21.28 & 22.39 \\
    & Ours & {\bf 5.24} & {\bf 7.89} & {\bf 8.24 } & {\bf 8.49 }  \\
    \midrule
    \multirow{2}{*}{InterCap} &CHORE & 7.12 & 12.59 &16.11 & 21.05\\
    & Ours & {\bf 6.76} & {\bf 10.32} & {\bf 9.35 } & {\bf 11.38 } \\
    \midrule
    \multirow{3}{*}{$\text{InterCap}^*$} & PHOSA & 11.20 & 20.57 & 24.16 & 43.06\\
    & CHORE & 7.01  & 12.81  & 16.10  & 21.08  \\
    & Ours & {\bf 6.78} & {\bf 10.34} & {\bf 9.35 } & {\bf 11.54 } \\
    \bottomrule[1.5pt]
    \end{tabular}
    \caption{Human and object tracking results on BEHAVE~\cite{bhatnagar22behave} and InterCap~\cite{huang2022intercap} datasets (unit: cm). * denotes key frames only. w is the temporal window size used for Procrustes alignment where w=1 means per-frame Procrustes and w=10 means alignment over a sliding window of 10s. We can see our method clearly outperforms baseline PHOSA\cite{zhang2020phosa} and CHORE\cite{xie22chore} in all metrics.}
    \label{table:main-results}
\end{table}

We compare our human and object tracking results against baseline PHOSA~\cite{zhang2020phosa} and CHORE~\cite{xie22chore} and report the errors in \cref{table:main-results}. Note that the comparison with PHOSA and CHORE is not strictly fair as they do not use temporal information. Nevertheless they are our closest baselines and we show that our method outperforms them using per-frame alignment and is significantly better under a more relevant evaluation metric (align w=10) for video tracking. We also show qualitative comparisons in \cref{fig:main-result}. It can be seen that both PHOSA and CHORE are not able to accurately capture the object under heavy occlusions while our method is more robust under these challenging occlusion cases.

\subsection{Importance of \ourSMPL{} conditioning}
We propose to condition our interaction field predictions on \ourSMPL{} meshes, we evaluate this for the joint tracking task in \cref{tab:eval-smpl-feat}. Our \ourSMPL{} conditioning allows us to obtain consistent relative translation instead of predicting the human at fixed depth. This significantly reduces the error when evaluating using alignment of temporal sliding window, see \cref{tab:eval-smpl-feat} and qualitative examples in Supp. 

Note that the \sifNet{} prediction relies on the estimated \ourSMPL{} discussed in \cref{sec:TranSMPL}. We also evaluate how the noisy \ourSMPL{} predictions can affect the joint tracking performance. We input GT SMPL to our \sifNet{} and \hvopNet{} to predict the object pose and perform joint optimization. The object errors (Chamfer distance in cm) are: 17.29 (w/o SMPL-T cond.), 8.23 (w/ SMPL-T cond., ours), 6.50 (GT SMPL). Our method is robust and is close to the ideal lower bound with GT SMPL.

\begin{table}[]
         
    \footnotesize
    \centering
    \begin{tabular}{c|c|c|c|c}
        Method & w/o \ourSMPL{}& w/o HVOP-Net &w/o joint opt. & \textbf{Ours}  \\
        \hline
         SMPL $\downarrow$ & 14.40 & 8.05 & 8.20 & \textbf{8.03} \\
         Obj. $\downarrow$ & 17.29 & 9.36 & 16.02 & \textbf{8.23}
    \end{tabular}
    \caption{Ablation studies. We report the joint tracking error (cm) after an alignment window of 10s. It can be seen that our proposed \ourSMPL{} conditioning, \hvopNet{} and joint optimization are important to achieve the best results. }
    
    \label{tab:eval-smpl-feat}
\end{table}

\subsection{Importance of \hvopNet{}}
\begin{figure}[t]
    \centering
    \includegraphics[width=0.46\textwidth]{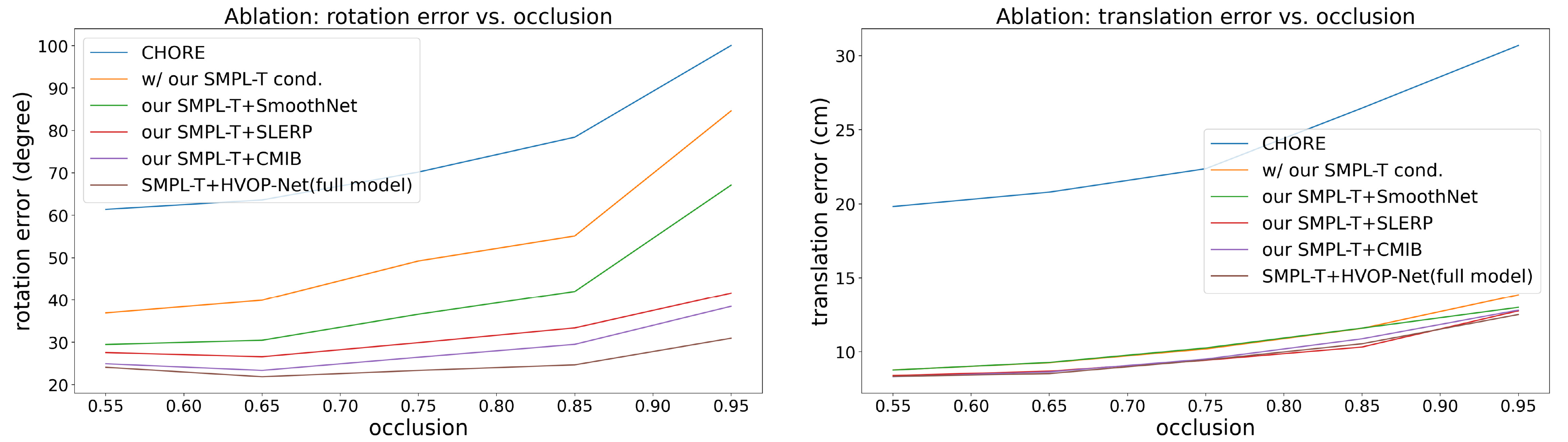}
    \caption{Object rotation(left) and translation(right) error vs. occlusion (1-fully occluded) for variants of our method. Our full model with \hvopNet{} predicts more robust rotation in occlusions.}
    \label{fig:error-vs-occlusion}
\end{figure}
\begin{figure}
    \centering
    \includegraphics[width=0.46\textwidth]{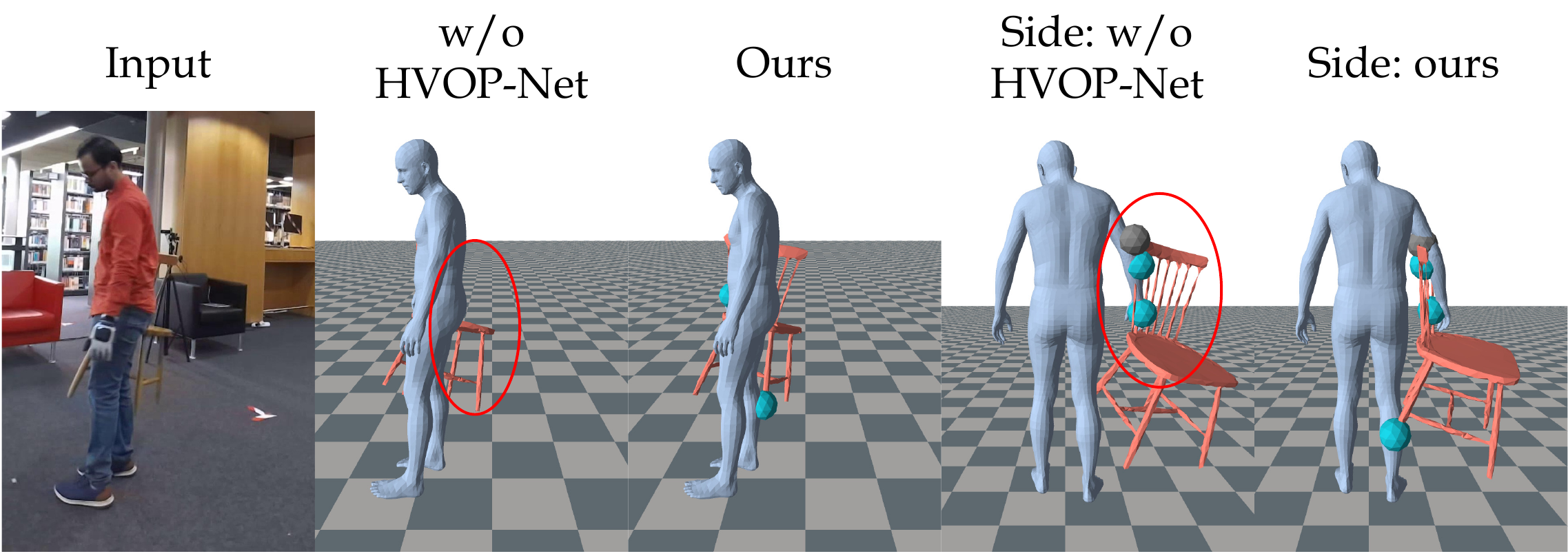}
    \caption{Importance of our \hvopNet{} object pose prediction. We can see that our \hvopNet{} corrects erroneous object pose under heavy occlusions. Better viewed after zoom in.}
    \label{fig:eval-pose-pred}
\end{figure}

We propose a \emph{human and visibility aware} object pose prediction network (\hvopNet{}) to reason about the object under heavy occlusions. Without this component, the object error is much higher (\cref{tab:eval-smpl-feat} column 3), which suggests the importance of our \hvopNet{}. 

For the object pose prediction task, there are other similar alternatives to \hvopNet{}:
1). Linear interpolation (SLERP), infill the object pose of invisible frames using spherical linear interpolation between two visible frames.
2). SmoothNet \cite{zeng2022smoothnet}, a fully connected neural network trained to smooth the object motion.  3). CMIB \cite{KIM2022108894CMIB}, a SoTA method for human motion infilling. To evaluate the effectiveness of our \hvopNet{}, we replace \hvopNet{} with these methods and run the full tracking pipeline respectively. The SMPL errors are similar (deviate $<$ 0.1cm) as \hvopNet{} only affects objects. We separate object error into rotation (angle distance) and translation, and further analyse the error under varying occlusion for CHORE, our method (\ourSMPL{} + \hvopNet{}) and its variants (\ourSMPL{} + SmoothNet/SLERP/CMIB) in Fig.~\ref{fig:error-vs-occlusion}. 

All methods except CHORE predict similar translation (due to \ourSMPL{} cond.) but our HVOP-Net obtains clearly more robust rotation under heavy occlusions. SmoothNet smooths the object motion but cannot correct errors from long-term occlusions. SLERP and CMIB \cite{KIM2022108894CMIB} are able to correct some pose errors but do not take the human context into account hence cannot handle heavy occlusions very well. Our method leverages the human motion and object pose from visible frames hence achieves the best result. We show one example where our \hvopNet{} corrects the erroneous raw prediction in \cref{fig:eval-pose-pred}. Please see our supplementary for more examples.

\subsection{Generalization}
\begin{figure}[t]
    \centering
    \includegraphics[width=0.46\textwidth]{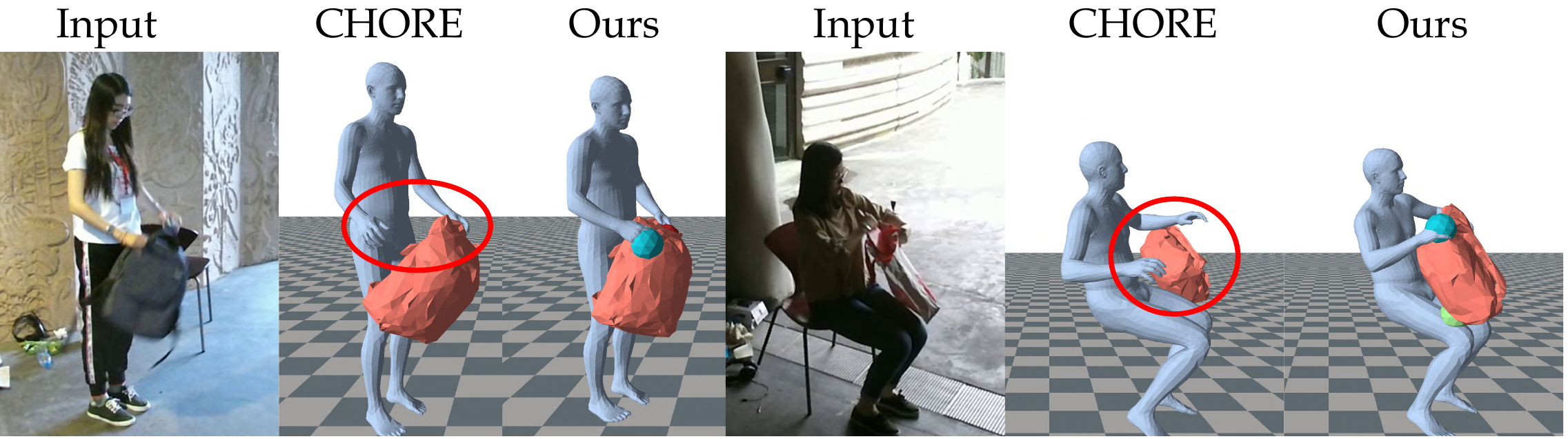}
    \caption{Results in NTU-RGBD dataset \cite{Liu_2019_NTURGBD120}. It can be seen that our method generalizes well and wormks better than CHORE\cite{xie22chore}. 
    }
    \label{fig:generalization}
\end{figure}
To verify the generalization ability of our method, we apply our model trained on BEHAVE to the NTU-RGBD~\cite{Liu_2019_NTURGBD120} dataset. We leverage Detectron~\cite{wu2019detectron2}, interactive~\cite{fbrs2020} and video~\cite{cheng2021mivos} segmentation to obtain the input human and object masks. 
Two example comparisons with CHORE~\cite{xie22chore} are shown in \cref{fig:generalization}. We can see our method generalizes well to NTU-RGBD and works better than CHORE. Please see Supp. for evaluation details and more comparisons.

\section{Limitations}
Although our method works robustly under heavy occlusions, there are still some limitations. For instance, we assume known object templates for tracking. An interesting direction is to build such a template from videos as demonstrated by recent works \cite{yang2021lasr, yang2021viser, yang2022banmo, wu2021dove}. Our method may fail under challenging cases such as significant object pose change under heavy occlusions. It can also make noisy pose predictions when the object is symmetric or the pose is uncommon. Example failure cases are shown in Supp.

\section{Conclusion}
We present a novel method for tracking human, object and realistic contacts between them from monocular RGB cameras. Our first contribution is a \ourSMPL{} conditioned neural field network that allows consistent and more accurate 3D reconstruction in a video sequence. Our second contribution is a human motion and object visibility aware network that can recover the object pose under heavy occlusions. Our experiments show that our method significantly outperforms state of the art methods in two datasets. Our extensive ablations show that our \ourSMPL{} conditioning and \hvopNet{} are important for accurate tracking. We also show that our method generalizes well to another dataset it is not trained on. Our code and model are released to promote future research in this direction.

\noindent
{\small
\textbf{Acknowledgements.} We thank RVH group members \cite{rvh_grp} for their helpful discussions. We also thank reviewers for their feedback which improves the manuscript. This work is funded by the Deutsche Forschungsgemeinschaft (DFG, German Research Foundation) - 409792180 (Emmy Noether Programme,
project: Real Virtual Humans), and German Federal Ministry of Education and Research (BMBF): Tübingen AI Center, FKZ: 01IS18039A. Gerard Pons-Moll is a Professor at the University of Tübingen endowed by the Carl Zeiss Foundation, at the Department of Computer Science and a member of the Machine Learning Cluster of Excellence, EXC number 2064/1 – Project number 390727645.

}

\begin{appendices}
\label{appendices}
In this supplementary, we first list all the implementation details of our method and then show more ablation study results as well as comparison with CHORE\cite{xie22chore} on NTU-RGBD\cite{Liu_2019_NTURGBD120} dataset. We end with discussions of failure cases and future works. 

\section{Implementation details}
\subsection{Obtaining \ourSMPL{} meshes}
To obtain the image-aligned SMPL meshes that have consistent translation (\ourSMPL{}) we keep the SMPL shape parameters and optimize the body pose and global translation values. The loss weights for this optimization are: $\lambda_\text{J2D}=0.09, \lambda_\text{reg}=1.0\times10^{-5}, \lambda_a=25, \lambda_\text{pi}=900$.  We optimize the parameters until convergence with a maximum iteration of 1000. 

\subsection{\sifNet{}: \ourSMPL{} conditioned interaction field}
\begin{figure}[t]
    \centering
    \includegraphics[width=0.46\textwidth]{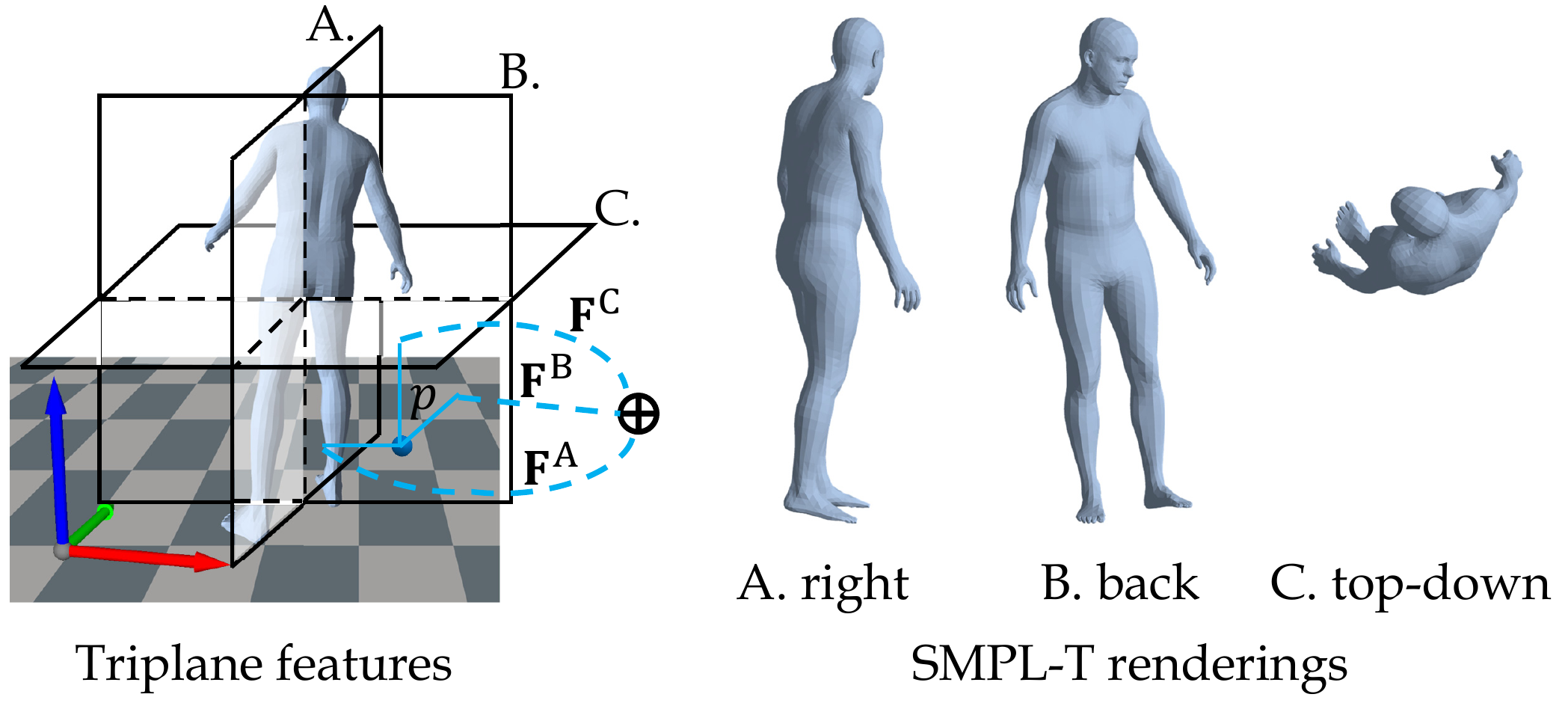}
    \caption{Visualization of our \ourSMPL{} triplane feature extraction and rendering. The triplane origin is placed at the \ourSMPL{} body center and we render the mesh from three views using orthographic projection: righ-left (A), back-front (B) and top-down (C). The query point $p$ is projected into the three planes using same projection for rendering and we extract pixel aligned features $\mat{F}^A, \mat{F}^B, \mat{F}^C$ from the feature planes respectively. Note that we render the \ourSMPL{} with color here for visualization, the actual input to our network are silhouette images only.}
    \label{fig:triplane-vis}
\end{figure}

A visualization of our \ourSMPL{} triplane rendering and query point projection can be found in \cref{fig:triplane-vis}. We discuss our network architecture and training details next. 

\paragraph{Network architecture.} We use the stacked hourglass network \cite{stacked-hourglass} for both RGB image encoder $f^\text{enc}$ and SMPL rendering encoder $f^\text{tri}$. We use 3 stacks for $f^\text{tri}$ and the output feature dimension is $d^\text{tri}_\text{o}=64$. Hence $f^\text{tri}:\mathbb{R}^{H\times W}\mapsto \mathbb{R}^{H/4\times W/4 \times 64}$ where $H=W=512$. We also use 3 stacks for $f^\text{enc}$ but the feature dimension is $d^\text{enc}_\text{o}=256$. Hence $f^\text{enc}:\mathbb{R}^{H\times W\times 5}\mapsto \mathbb{R}^{H/4\times W/4 \times 256}$. We also concatenate the image features extracted from the first convolution layer and query point coordinate to the features. Thus the total feature dimension to our decoders is: $d=(d^\text{tri}_1 + d^\text{tri}_\text{o})\times 3 + d^\text{enc}_1 + d^\text{enc}_\text{o} + 3=611$, here $d^\text{tri}_1=32, d^\text{enc}_1=64$. 
All decoders consist of three FC layers with ReLU activation and one output FC layer with hidden dimension of 128 for the intermediate features. The visibility decoder $f^v$ additionally has a sigmoid output activation layer. The output shape is $2, 14, 9, 3, 1$ for $f^u, f^p, f^R, f^c, f^v$ respectively. 

\paragraph{Training.} All feature encoders and decoders are trained end to end with the loss: $L=\lambda_u(L_{u_h}+L_{u_o}) + \lambda_p L_p +\lambda_R L_R + \lambda_c L_c + \lambda_v L_v$. Here $L_{u_i}$ is the $L_1$ distance between ground truth and predicted unsigned distance to human or object surface \cite{xie22chore}. $L_p$ is a standard categorical cross entropy loss for SMPL part correspondence prediction. $L_R, L_c, L_v$ are mean square losses between ground truth and predicted values for rotation matrix, translation vector and visibility score respectively. The loss weights are: $\lambda_u=1.0, \lambda_R=0.006, \lambda_c=500, \lambda_c=\lambda_v=1000$. The model is trained for 18 epochs and it takes 25h to converge on a machine with 4 RTX8000 GPUs each with 48GB memory. The training batch size is 8. 

\subsection{\hvopNet{}: object pose under occlusion}
We use three transformers $f^s, f^o, f^\text{comb}$ to aggregate features from \ourSMPL{}, object pose and joint human object information respectively. We use the 6D vector \cite{Zhou_2019_CVPR_rot6d} to represent the ration matrix of \ourSMPL{} and object pose parameters. Hence the \ourSMPL{} pose dimension is $24\times6+3=147$, where 3 denotes the global translation. We predict the object rotation only thus the object data dimension is 6. The \ourSMPL{} transformer $f^s$ consists of an MLP: $\mathbb{R}^{T\times 147}\mapsto \mathbb{R}^{T\times 128}$ and two layers of multi-head self-attention (MHSA) module \cite{NIPS2017_attention} with 4 heads. Similarly, the object transformer $f^o$ consists of an MLP: $\mathbb{R}^{T\times 6}\mapsto \mathbb{R}^{T\times 32}$ and two layers of MHSA module with 2 heads. The joint transformer $f^\text{comb}$ consists of 4 layers of MHSA module with 1 head only. GeLU activation is used in all MHSA modules. We finally predict the object pose using two MLP layers with an intermediate feature dimension of 32 and LeakyReLU activation. 

The model is trained to minimize the $L_1$ losses of pose value and accelerations: $L=\lambda_\text{pose}L_\text{pose} + \lambda_\text{accel}L_\text{accel}$, where $\lambda_\text{pose}=1.0, \lambda_\text{accel}=0.1$. It is trained on a server with 2 RTX8000 GPUs, each GPU has 48GB memory capacity. It takes around 7h to converge (64 epochs).

\subsection{SmoothNet for \ourSMPL{} and object}
We use SmoothNet\cite{zeng2022smoothnet} to smooth our \ourSMPL{} and \sifNet{} object pose predictions. We use exactly the same model and training strategy proposed by the original paper. The input to the \ourSMPL{} SmoothNet is our estimated \ourSMPL{} pose and translation (relative to the first frame). The input to the object SmoothNet is the object rotation (6D vector). Following the standard practice of SmoothNet\cite{zeng2022smoothnet}, we train both models on the predictions from the BEHAVE \cite{bhatnagar22behave} training set. Note that we do not fine-tune them on InterCap\cite{huang2022intercap} dataset. We evaluate this component in \cref{sec:eval-smoothnet}. 

\subsection{Visibility aware joint optimization}
The objective function defined in Eq. 2 is highly non-convex thus we solve this optimization problem in two stages. We first optimize the SMPL pose and shape parameters using human data term only. We then optimize the object parameters using the object and contact data terms. The loss weights are set to: $\lambda_\text{reg}=2.5\times 10^{-4}, \lambda_\text{ah}=10^4, \lambda_h=10^4, \lambda_p=t\times 10^{-4}, \lambda_o=900, \lambda_\text{occ}=9\times 10^{-4}, \lambda_\text{ao}=225, \lambda_c=900$, where $\lambda_c$ is the loss weight for the contact data term defined in Eq. 5.

\section{Additional ablation results}
\subsection{Further evaluation of \ourSMPL{} conditioning} \label{supp-sec:eval-smplt}
\begin{figure}[t]
    \centering
    \includegraphics[width=0.46\textwidth]{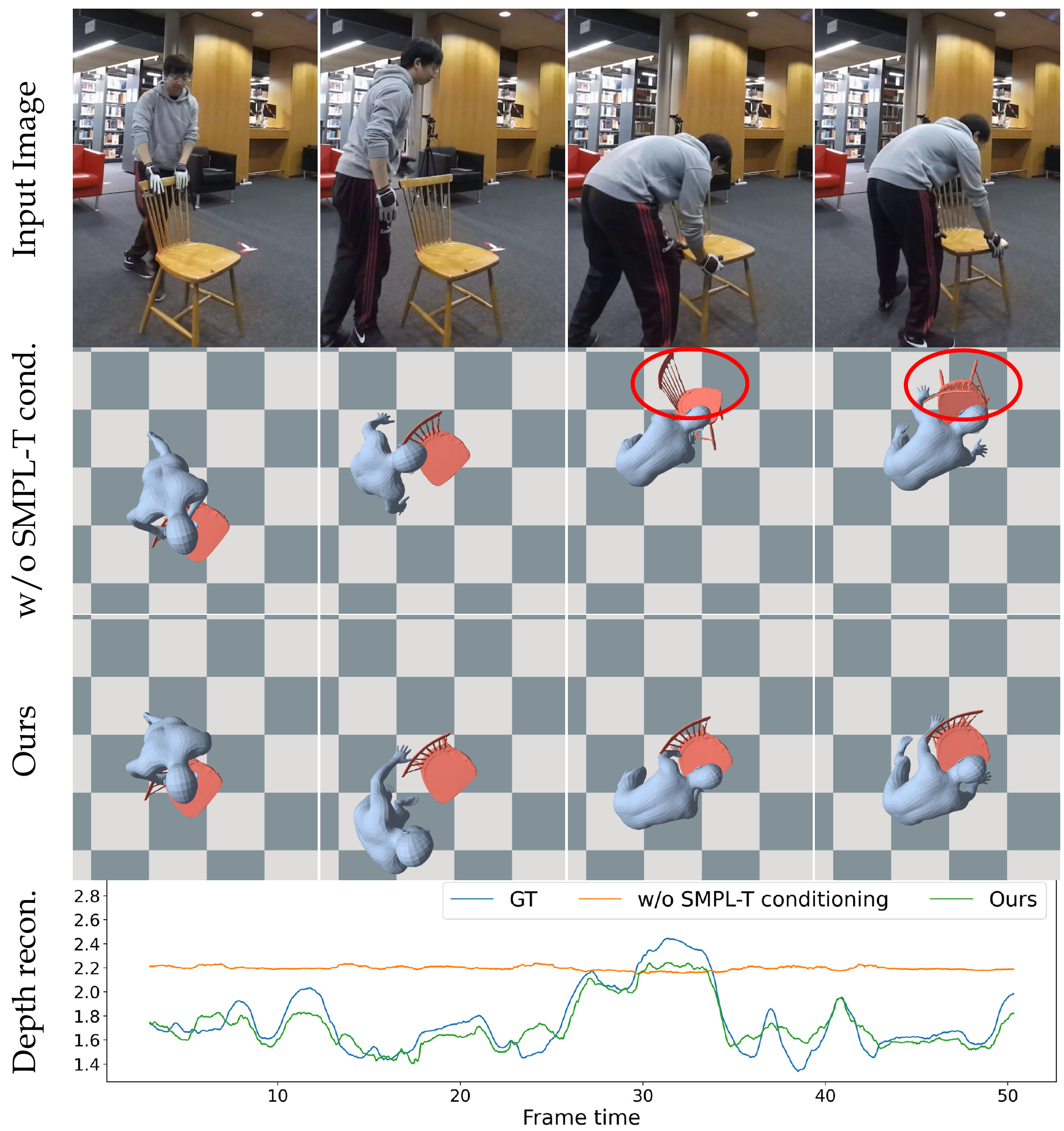}
    \caption{Evaluating \ourSMPL{} conditioning for neural field prediction. We can see that without conditioning on \ourSMPL{} meshes, the object pose prediction is worse and human is reconstructed at fixed depth, leading to inconsistent relative location across frames. Our method recovers the relative translation more faithfully and obtain better object pose predictions. 
    }
    \label{fig:eval-smpl-feature}
\end{figure}

We show some example images from one sequence in \cref{fig:eval-smpl-feature} to evaluate the importance of our \ourSMPL{} conditioning. It can be seen that without this conditioning, the human is reconstructed at fixed depth, leading to inconsistent relative translation across time. Our method predicts more coherent relative human translation and more accurate object pose. 

To further evaluate \ourSMPL{} conditioning, we compute the object pose error from the raw network predictions and compare it with the object pose of CHORE which is also the raw prediction from the network. The pose error is computed as Chamfer distance (CD) and vertex to vertex (v2v) error after centring the prediction and GT mesh at origin. We also report the translation error (transl.) as the distance between predicted and GT translation. The results are shown in \cref{tab:eval-smplt-feat}. We can clearly see that our SMPL feature improves both the raw object pose prediction and distance fields (results after optimization are also improved). 

\begin{table}[h]
    \small %
    \centering
    \begin{tabular}{ l| c c c c c}
    \hline
         \multirow{2}{*}{Method} & \multicolumn{3}{c}{Raw prediction} & \multicolumn{2}{c}{After opt. w=10}\\
         
          & CD$\downarrow$ & v2v$\downarrow$ & transl.$\downarrow$ & SMPL$\downarrow$ &obj.$\downarrow$\\
          \hline
         w/o \ourSMPL{} & 5.56 & 16.10 & 14.28 & 14.40 & 17.29 \\
         Ours & {\bf 3.98} & {\bf 12.34} & {\bf 9.53} & {\bf 8.03} & {\bf 8.23}\\
         \hline
    \end{tabular}
    \caption{Importance of \ourSMPL{} conditioning (errors in cm). We can see that our \ourSMPL{} feature improves both the raw object pose prediction and distance fields (after opt.). Without our \ourSMPL{} conditioning, the reconstructed translation is not consistent across frames, leading to large errors after alignment of temporal window of 10s (w=10).}
    \label{tab:eval-smplt-feat}
\end{table}

\subsection{Comparing different pose prediction methods}
\begin{figure*}[t]
    \centering
    \includegraphics[width=0.98\textwidth]{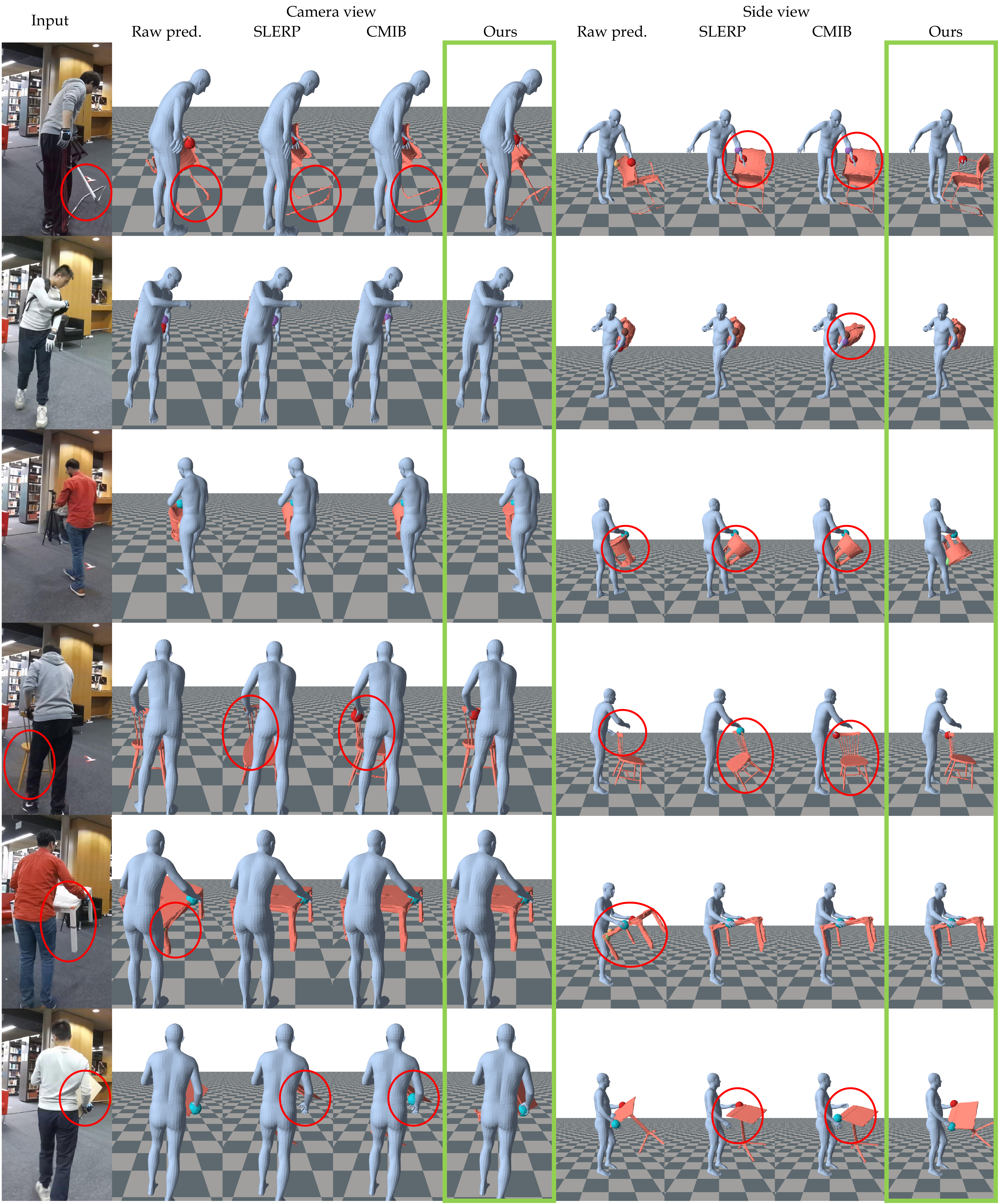}
    \caption{Comparing different object pose prediction method under heavy occlusions. Raw prediction is from our \sifNet{} output, SLERP denotes linear interpolation and CMIB is from \cite{KIM2022108894CMIB}. We can see SLERP and CMIB can correct some errors (row 5) but they do not take the human motion into account hence often fail in more challenging cases. Our method is more robust as it leverages information from both human motion and object pose from visible frames.}
    \label{fig:comp-pose-preds}
\end{figure*}

We show some example comparisons of different object pose prediction methods under heavy occlusions in \cref{fig:comp-pose-preds}. We compare our method against: 1). Raw prediction from our \sifNet{}. 2). Linearly interpolate the occluded poses from visible frames (SLERP). 3). CMIB \cite{KIM2022108894CMIB}, a transformer based model trained to infill the object motion using visible frames. Note here the evaluation is based on the final tracking results and we report the object errors only as the difference of SMPL error is very small. Similar to \cref{supp-sec:eval-smplt}, the object errors are computed as Chamfer distance, v2v error and translation error. 

It can be seen that the raw pose prediction is noisy due to occlusion. SLERP and CMIB corrects some pose errors but is not robust as they do not leverage the human information. Our method is more accurate as it takes the human context and object pose into account.

\subsection{Evaluating SmoothNet}\label{sec:eval-smoothnet}
\begin{table}
    \small  
    \centering
    \begin{tabular}{l|c c c}
    \hline
        Method & Chamfer & v2v & Acceleration \\
        \hline
        w/o SmoothNet & 8.71 & 9.84 & 1.38\\
        w/ SmoothNet & {\bf 8.01} & {\bf 9.12} & {\bf 1.18}\\
        \hline
    \end{tabular}
    \caption{Ablate SMPL SmoothNet (errors in cm). We can see that SmoothNet ~\cite{zeng2022smoothnet} improves the overall smoothness and slightly reduces the pose errors.}
    \label{tab:eval-smoothnet-smpl}
\end{table}

\begin{table}
    \small  
    \centering
    \begin{tabular}{l|c c c}
    \hline
        Method & Chamfer & v2v & Translation \\
        \hline
        a. Raw prediction & 5.03 & 10.39 & 10.01 \\
        b. Raw + SmoothNet & 4.22 & 8.60 & 10.16 \\
        c. Raw + our pose pred. & 4.09 & 8.02 & 10.20\\
        d. Our full model & {\bf 3.62 } & {\bf 7.20} & {\bf 9.96} \\
        \hline
    \end{tabular}
    \caption{Ablate SmoothNet for object pose prediction (errors in cm). We can see our pose prediction (c) is better than SmoothNet~\cite{zeng2022smoothnet} (b). Combing both we obtain the best result (d).}
    \label{tab:eval-smoothnet-object}
\end{table}
SmoothNet \cite{zeng2022smoothnet} is used to smooth the \ourSMPL{} parameters after 2D keypoint based optimization. We evaluate this step by computing the SMPL errors, shown in \cref{tab:eval-smoothnet-smpl}. We can see that SmoothNet reduces the SMPL error slightly. 

We also use SmoothNet to smooth the object pose before sending it to our human and visibility aware object pose prediction network. SmoothNet cannot correct errors under long-term occlusions. However, it provides smoother object motion for visible frames which can benefit our pose prediction network. We evaluate this using object pose errors and report the results in \cref{tab:eval-smoothnet-object}. It can be seen that our method (\cref{tab:eval-smoothnet-object}c) works better than SmoothNet (\cref{tab:eval-smoothnet-object}b) on raw predictions. Nevertheless, with smoothed pose after SmoothNet, our method achieves the best result (\cref{tab:eval-smoothnet-object} d).

\subsection{Runtime cost}
\ourSMPL{} pre-fitting and joint optimization can be run in batches hence the average runtime per frame is not long: \ourSMPL{} pre-fitting: 6.38s, SIF-Net object pose prediction: 0.89s, HVOP-Net: 1.3ms, joint optimization: 9.26s, total: 16.53s. Compared to CHORE ($\sim$12s/frame)~\cite{xie22chore}, the additional cost is mainly from the \ourSMPL{} pre-fitting. Yet, \ourSMPL{} conditioning allows faster convergence of joint optimization and much better reconstruction. Since we use efficient 2D encoder instead of 3D encoder, it takes only 1.05GB GPU memory to load the SIF-Net model. This allows us to do joint optimization with batch size up to 128 on a GPU with 48GB memory. 

\section{Generalization to NTU-RGBD dataset}
\paragraph{Obtaining input masks.} Unlike BEHAVE and InterCap where the human and object masks are provided by the dataset, there are no masks in NTU-RGBD. To this end, we run DetectronV2~\cite{wu2019detectron2} to obtain the human masks. We manually segment the object in the first frame using interactive segmentation~\cite{fbrs2020} ($<$1min/image) and then use video segmentation ~\cite{cheng2021mivos} to propagate the masks. The overhead of 1min/video manual label is small. 

\begin{figure*}
    \centering
    \includegraphics[width=0.98\textwidth]{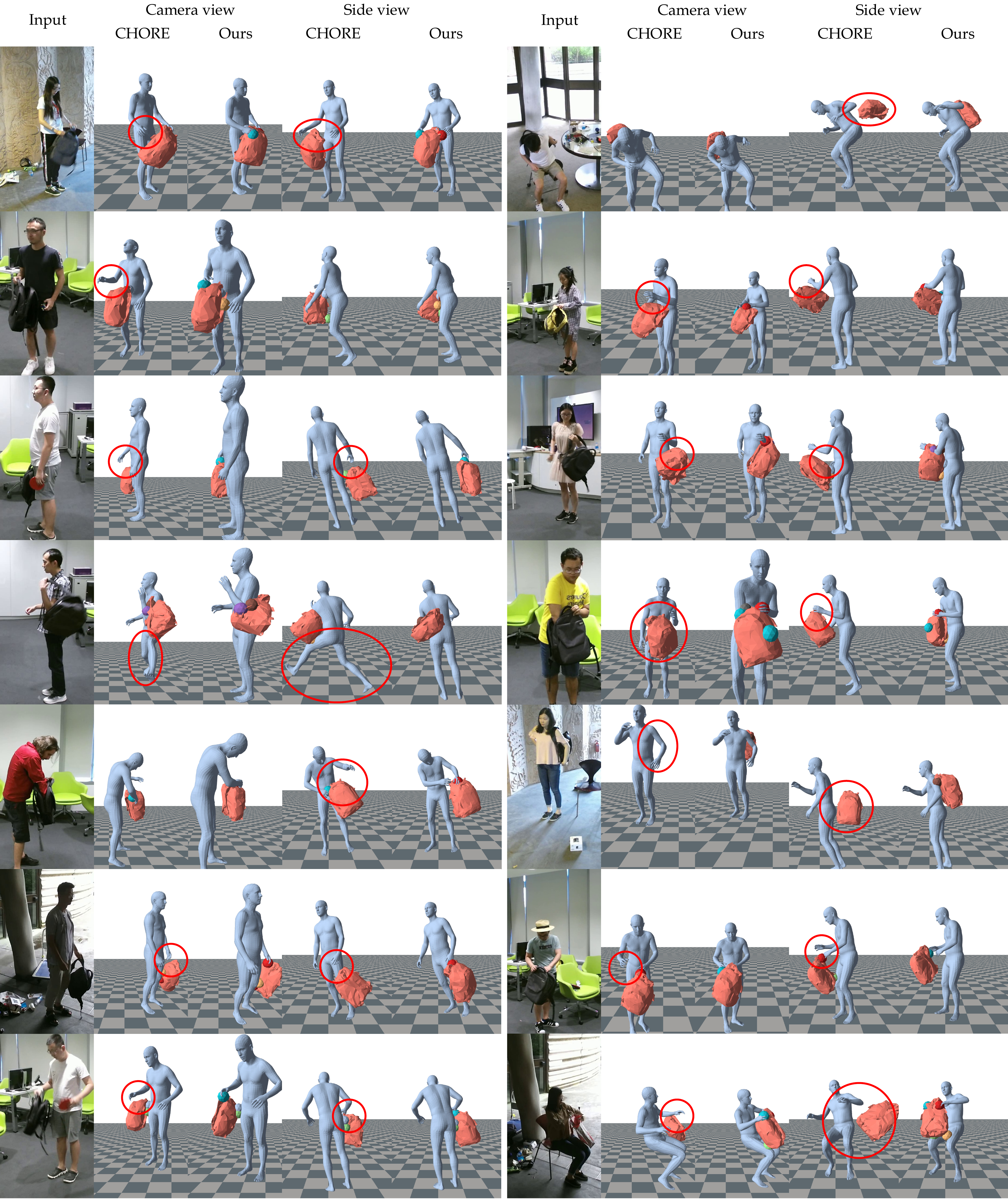}
    \caption{Comparing our method with CHORE\cite{xie22chore} on NTU-RGBD \cite{Liu_2019_NTURGBD120} dataset. It can be seen that CHORE does not capture the realistic contacts between the person and the backpack. Our method recovers the 3D human, the object and contacts more faithfully in different interaction types, camera view points and backgrounds.}
    \label{fig:comp-ntu}
\end{figure*}

We show more results from our method on NTU-RGBD dataset\cite{Liu_2019_NTURGBD120} and compare against CHORE\cite{xie22chore} in \cref{fig:comp-ntu}. It can be seen that CHORE may predict some reasonable object pose but it fails quite often to capture the fine-grained contacts between the human and object. Our method obtains more coherent reconstruction for different subjects, human-backpack interactions, camera view points and backgrounds. Please see our \href{https://virtualhumans.mpi-inf.mpg.de/VisTracker}{project website} for comparison in full sequences. 

\section{Limitations and future works}
\begin{figure}[t]
    \centering
    \includegraphics[width=0.46\textwidth]{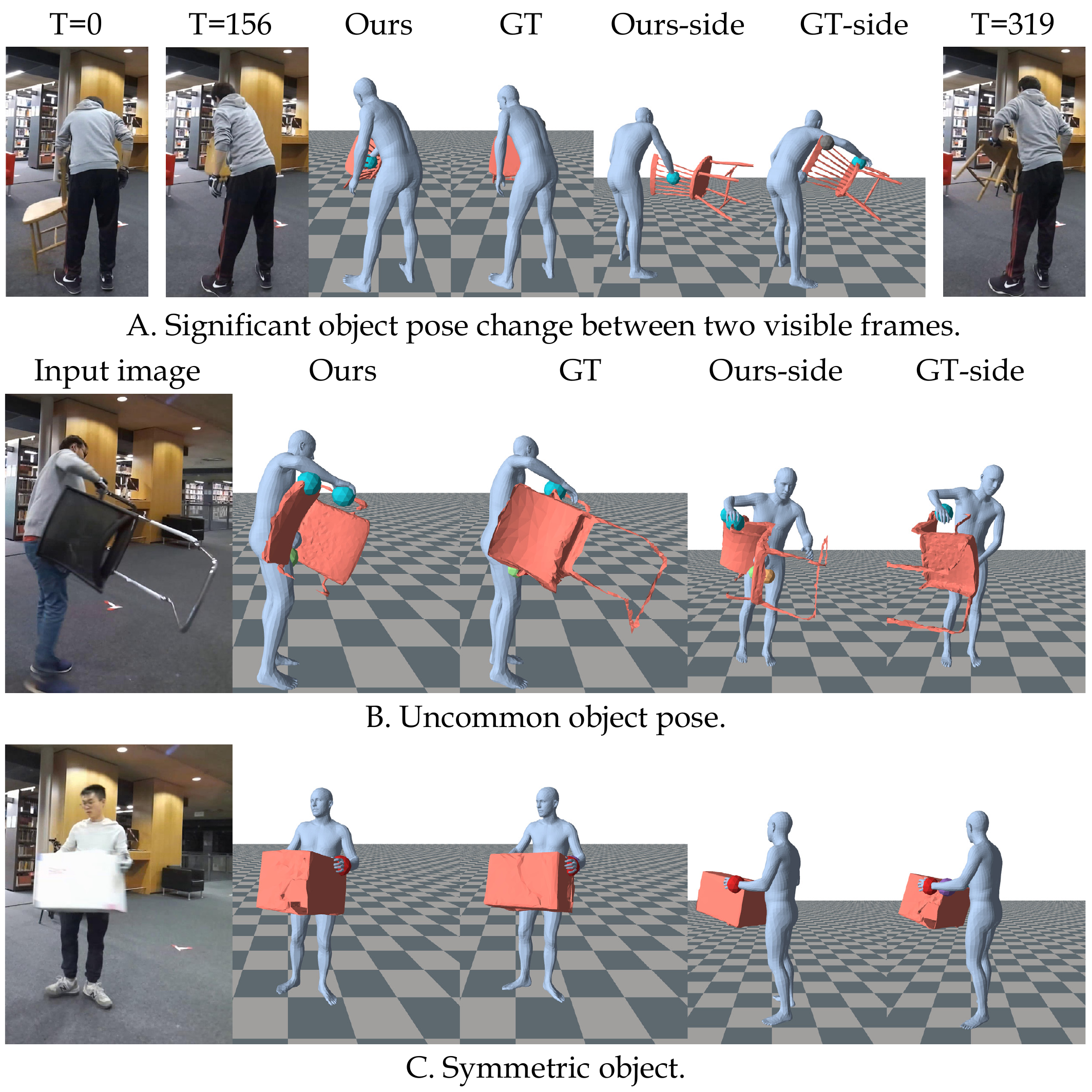}
    \caption{Failure cases analysis. We show three typical failure cases of our method: A. The occluded object pose (T=156) changes significantly between two visible frames (T=0 and T=319) and it is difficult to accurate track the contact changes. B. The object pose is not commonly seen during interaction and it is difficult to predict for this rare pose. C. The object is symmetric. The joint optimization satisfies the object mask and contacts but is not semantically correct.}
    \label{fig:failure-cases}
\end{figure}
Although our method works robustly under heavy occlusions, there are still some limitations. Firstly, we assume known object templates for tracking, an interesting direction is to build such a template from videos as demonstrated by recent works \cite{yang2021lasr, yang2021viser, yang2022banmo, wu2021dove}. Secondly, it would be interesting to model multi-person or even multi-object interactions which is a more realistic setting in real-life applications. In addition, the backpack can also deform non-rigidly which is not modelled in our method. Further works can incorporate the surface deformation \cite{Li_3DV2022MocapDeform} or object articulation \cite{xu2021d3dhoi} into the human object interaction. We leave these for future works.   

We identify three typical failure cases of our method, some examples are shown in \cref{fig:failure-cases}. The first typical failure case comes from heavy occlusion when the object undergoes significant changes (object pose and contact locations) between two visible frames. In this case, it is very difficult to track the pose and contact changes accurately (\cref{fig:failure-cases} A). Second typical failure is due to the difficulty of pose prediction itself even the object is fully visible. In this case the object pose is uncommon and the network failed to predict it correctly (\cref{fig:failure-cases} B). Another failure is caused by symmetric objects. Our optimization minimizes the 2D mask loss and contact constraints but the network is confused by the symmetry and the initial pose prediction is not semantically correct (\cref{fig:failure-cases} C). In addition, the training data for these objects is very limited (only $1/3$ of other objects). More training data or explicitly reasoning about the symmetry \cite{zheng_tp-ae_2022} can be helpful.

\end{appendices}
{\small
\bibliographystyle{ieee_fullname}
\bibliography{egbib}
}

\end{document}